\documentclass[runningheads]{llncs}
\usepackage{fontawesome5}

\usepackage[T1]{fontenc}
\usepackage{graphicx}
\usepackage{amssymb}
\usepackage{mathrsfs}
\usepackage{newtxmath}
\usepackage{array} 
\usepackage{multirow}
\usepackage{ragged2e}
\usepackage[colorlinks,linkcolor=blue]{hyperref}
\usepackage{color}
\begin{document}
	\title{Pick the Best Pre-trained Model: Towards Transferability Estimation for Medical Image Segmentation}

	\author{Yuncheng Yang\inst{1} \and
		Meng Wei\inst{3} \and
		Junjun He\inst{3}
		\and Jie Yang\inst{1}\faIcon{envelope} \and Jin Ye\inst{3} \and Yun Gu\inst{1, 2}\faIcon{envelope}}
	\authorrunning{Y. Yang et al.}
	%
	\institute{
		Institute of Image Processing and Pattern Recognition, Shanghai Jiao Tong University, Shanghai, China \and
		Institute of Medical Robotics, Shanghai Jiao Tong University, Shanghai, China \and Shanghai AI Lab, Shanghai, China}

	\maketitle              
	\begin{abstract}
		Transfer learning is a critical technique in training deep neural networks for the challenging medical image segmentation task that requires enormous resources. With the abundance of medical image data, many research institutions release models trained on various datasets that can form a huge pool of candidate source models to choose from. Hence, it’s vital to estimate the source models’ transferability (i.e., the ability to generalize across different downstream tasks) for proper and efficient model reuse. To make up for its deficiency when applying transfer learning to medical image segmentation, in this paper, we therefore propose a new  \textbf{Transferability Estimation} (TE) method. We first analyze the drawbacks of using the existing TE algorithms for medical image segmentation and then design a source-free TE framework that considers both class consistency and feature variety for better estimation. Extensive experiments show that our method surpasses all current algorithms for transferability estimation in medical image segmentation. Code is available at \href{https://github.com/EndoluminalSurgicalVision-IMR/CCFV}{here}.

	\end{abstract}
	
	\section{Introduction}
	The development of deep neural networks has greatly promoted medical imaging-based computer-aided diagnosis. Due to the large amount of learnable parameters in neural networks, sufficient annotated training samples are required for training. However, the labeling process of medical images is tedious and time-consuming. To address this problem, the common paradigm of \textit{transfer learning}, which first pre-trains a model on upstream image datasets and then fine-tunes it on various target tasks, has been widely investigated in recent years~\cite{tajbakhsh2016convolutional,zhou2021active,irvin2019chexpert}. Compared with the distributed training across multiple centers, there are no specific ethical issues or computational design of distributed/federated learning frameworks with the ``pre-train-then-fine-tune'' workflow.
	
	Previous works mainly focused on the fine-tuning strategy to effectively adapt the knowledge from the pre-trained models to target tasks~\cite{xuhong2018explicit,li2019delta,chen2019bss,reiss2021panda}. With the increasing number of pre-trained networks provided by the community, model repositories like Hugging Face\cite{wolf2020transformers} and PyTorch Hub\cite{paszke2019pytorch} enable researchers to experiment across a large number of downstream datasets and tasks. These pre-trained models require less training time and have better performance and robustness compared with the learning-from-scratch models. However, it has been observed by recent works~\cite{wang2020understanding} that the pre-trained models cannot always benefit the downstream tasks.  When the knowledge is transferred from a less relevant source, it may not improve the performance or even negatively affect the intended outcome~\cite{wang2019characterizing}. A brute-force method is to fine-tune a set of pre-trained models with target datasets to find the optimal one. This process is time-consuming and laborious. Existing methods also measured the task-relatedness between source and target datasets\cite{dwivedi2020duality,dwivedi2019representation,zamir2018taskonomy,tong2021mathematical}. However, most of these works require source information available while medical images have more privacy and ethical issues and fewer datasets are publicly available than natural images.
	
	\begin{figure}[!t]
		\centering
		\includegraphics[width=1.0\linewidth]{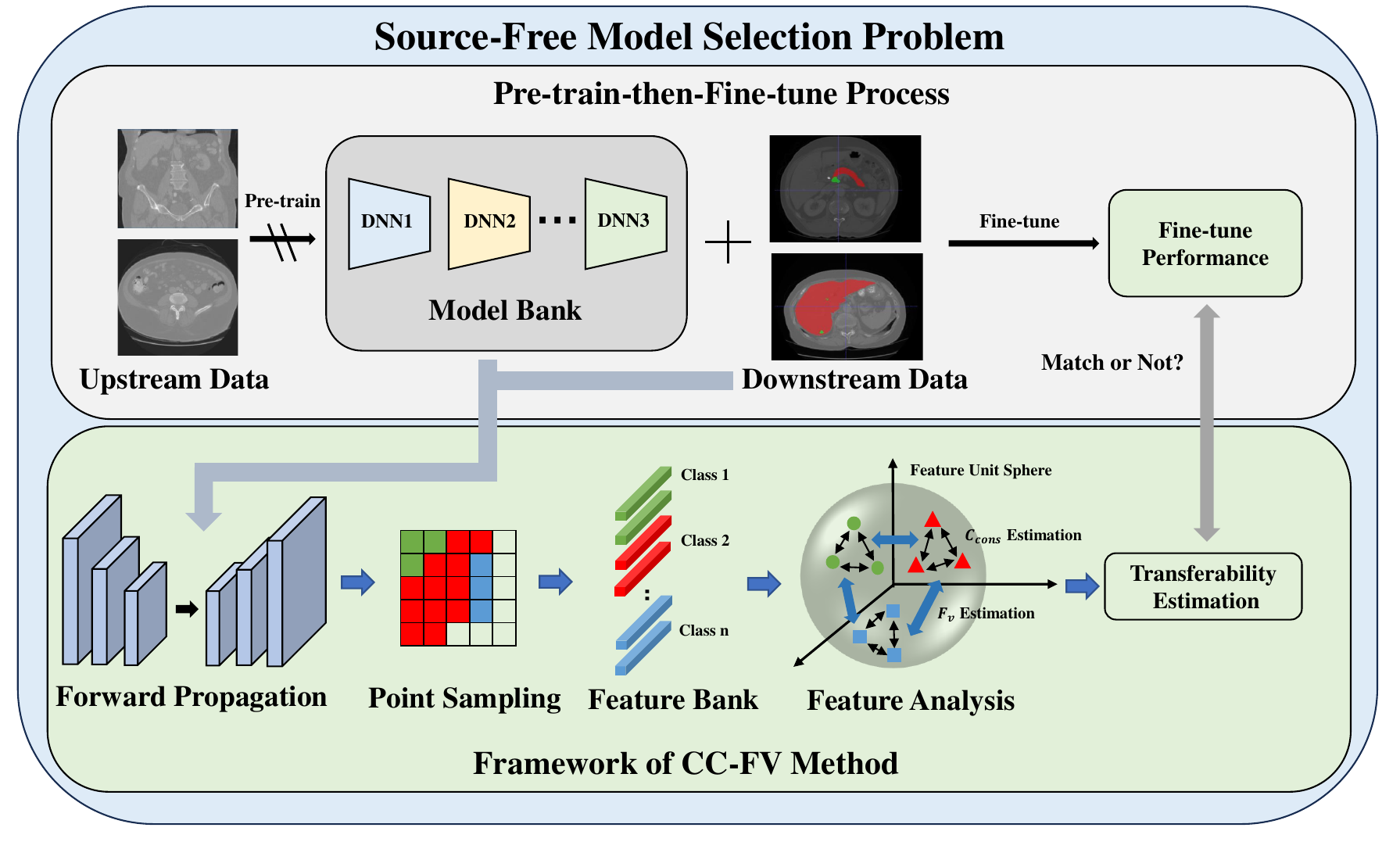}
		\caption{Source-free model selection problem and the framework of our \textbf{C}lass \textbf{C}onsistency with \textbf{F}eature \textbf{V}ariety constraint(CC-FV)  TE method. Our main goal is to predict the performance of models in the model bank after fine-tuning on downstream tasks without actually fine-tuning. Note that the upstream data are not available in our model selection process.}
		\label{fig: framework}
	\end{figure}
	
	Considering the issues mentioned above, this work focused on source-free pre-trained model selection for segmentation tasks in the medical image. As shown in Fig.\ref{fig: framework}, models pre-trained by upstream data constitute the model bank. The main idea is to directly measure the transferability of the pre-trained models without fully training based on the downstream/target dataset. Among the recent works, LEEP\cite{nguyen2020leep} and its variant\cite{agostinelli2022transferability,li2021ranking} were developed to utilize the log-likelihood between the target labels and the predictions from the source model. LogME~\cite{you2021logme} computed evidence based on the linear parameters assumption and efficiently leverages the compatibility between features and labels. GBC~\cite{pandy2022transferability} applied the Gaussian distribution to each class, and estimate the separability between classes as the basis for transferability estimation. TransRate\cite{huang2022frustratingly} evaluated the transferability of models with the compactness and the completeness of embedding space. Cui et.al~\cite{cui2022discriminability} contended that discriminability and transferability are crucial properties of representations and introduce the information bottleneck theory for transferability estimation. These methods have achieved promising performance on classification and regression tasks without fully considering the properties of medical image segmentation. First, unlike classification and regression problems that can use a single n-dimensional feature vector to represent each image, segmentation problems lack a global semantic representation, which poses difficulties for direct transferability estimation. In addition, most label-comparison-based methods\cite{nguyen2020leep,huang2022frustratingly,you2021logme,pandy2022transferability} focus on the relationship between the embeddings and downstream labels without exploring the effectiveness of the features themselves. Third, medical images face severe class imbalance problems, with excessive differences between foreground and background. However, existing algorithms rarely give additional attention to the class imbalance problem. Besides, for semantic segmentation tasks, the feature pyramid is critical for the segmentation output of multi-scale objects while existing works neglect it. In our work, we propose a new method using class consistency and feature variety(CC-FV) with an efficient framework to estimate the transferability in medical image segmentation tasks. Class consistency employs the distribution of features extracted from foreground voxels of the same category in each sample to model and calculate their distance, the smaller the distance the better the result; feature diversity utilizes features sampled in the whole global feature map, and the uniformity of the feature distribution obtained by sampling is used to measure the effectiveness of the features themselves. Extensive experiments have proved the superiority of our method compared with baseline methods.

	\section{Methodology}
	
	\subsection{Problem Formulation}
	In our work, a model bank $\mathbb{M}$ consisting of pre-trained models $\{M_i\}_{i=1}^{K}$ are available to be fine-tuned and evaluated with a target dataset $\mathbb{D} = \{X_j, Y_j\}_{j=1}^{N}$, where $X_j$ is the image and $Y_j$ is the ground truth of segmentation. After fine-tuning, the performance of $M_i$ can be measured with the segmentation metric (e.g. Dice score), which is denoted by $\mathcal{P}^{i}_{s\rightarrow t}$ in this paper. Our work is to directly estimate the transferability score $\mathcal{T}^{i}_{s\rightarrow t}$ without fine-tuning the model on target datasets. A perfect transferability score should preserve the ordering, i.e. $\mathcal{T}^{i}_{s\rightarrow t}>\mathcal{T}^{j}_{s\rightarrow t}$ if and only if $\mathcal{P}^{i}_{s\rightarrow t}>\mathcal{P}^{j}_{s\rightarrow t}$. 
	
	\subsection{Class Consistency with Feature Variety Constraint TE method}
	The transferability of models from a weakly related source domain to a target domain can be compromised if the domains are not sufficiently comparable~\cite{wang2019characterizing}.
	This intrigues us about the question of "what kind of models are transferable". The proposed method is intuitive and straightforward: features extracted by the pre-trained model should be consistent within the class of the target dataset while representative and various globally. Therefore, \textbf{C}lass \textbf{C}onsistency and \textbf{F}eature \textbf{V}ariety are considered to estimate the transferability between models and downstream data.
	
	\noindent\textbf{Class Consistency.}
	The pre-trained models are trained with specific pretext tasks based on the upstream dataset. Therefore, features extracted by the pre-trained models cannot perfectly distinguish the foreground and background of target data. If the features are generalizable, foreground region features will likely follow a similar distribution even without fine-tuning. 
	
	Given a pair of target data $X_j$ and $X_{j'}$, the distribution of the features is modeled with the $n$-dimensional Gaussian distribution. Since the size of the foreground class varies across the cases, we therefore randomly sample the pixels/voxels of $X_j$ and $X_{j'}$ for each class and establish the feature distribution $F_j^{k}$, $F_{j'}^{k}$ based on the voxels of the \textit{k}th class to approximate the case-wise distribution of different classes. The class consistency between the data pair is measured by the Wasserstein distance~\cite{panaretos2019statistical} as follows:
	\begin{equation}
		\mathcal{W}_2^2(F_j^{k}, F_{j'}^{k})= \left\|\boldsymbol{\mu}_{F_j^{k}}-\boldsymbol{\mu}_{F_{j'}^{k}}\right\|^2+\operatorname{Tr}\left(\boldsymbol{\Sigma}_{F_j^{k}}\right)+\operatorname{Tr}\left(\boldsymbol{\Sigma}_{F_{j'}^{k}}\right)-2 \operatorname{Tr}\left(\left(\boldsymbol{\Sigma}_{F_j^{k}} \boldsymbol{\Sigma}_{F_{j'}^{k}}\right)^{1 / 2}\right)    
	\end{equation}
	where ${\mu}_{F_j^{k}}$, ${\mu}_{F_{j'}^{k}}$ are the mean of Gaussian distribution $F_j^{k}$, $F_{j'}^{k}$ and $\Sigma_{F_j^{k}}$ and $\Sigma_{F_{j'}^{k}}$ are covariance matrices of ${F_j^{k}}$ and ${F_{j'}^{k}}$. Compared to some commonly used metrics like KL-divergence or Bhattacharyya distance\cite{pandy2022transferability}, Wasserstein distance is more stable during the computation of high-dimensional matrices because it is unnecessary to compute the determinant or inverse of a high-dimensional matrix, which can easily lead to an overflow in numerical computation. We calculate the Wasserstein distance of the distribution with voxels of the same class in a sample pair comprised of every two samples in the dataset, and obtained the following definition of class consistency $C_{cons}$
	\begin{equation}
		\boldsymbol{C_{cons}} = \frac{1}{N(N-1)}{\sum_{k= 1}^{C}\sum_{i\ne j} \mathcal{W}_2(\boldsymbol{F_i^{k}}, \boldsymbol{F_j^{k}})}
	\end{equation}
	
	Given that 3D medical images are computationally intensive, and prone to causing out-of-memory problems, in the sliding window inference process for each case, we do not concatenate the output of each patch into the final prediction result, but directly sample from the patched output and concatenate them into the final sampled feature matrix. In the calculation of class consistency, we only sample the foreground voxels with a pre-defined sampling number which is proportional to the voxel number of each class in the image because of the severe class imbalance problem.
	
	\noindent\textbf{Feature Variety.} Class consistency is not the only criterion for transferability estimation. As a result of learning some trivial solutions, some overfitted models have limited generalization capacity and are difficult to apply to new tasks. We believe that the essential reason for this phenomenon is that class consistency is only concerned with local homogeneity of information while neglecting the integral feature quality assessment. Hence we propose the feature variety constraint, which measures the expressiveness of the features themselves and the uniformity of their probability distribution. Highly complex features are not easily overfitted in the downstream tasks and do not collapse to cause a trivial solution.
	
	To calculate the variety of features we need to analytically measure the properties of the feature distribution over the full feature space. Besides, to prevent overfitting and trivial features, we expect the distribution of features in the feature space to be as uniform and dispersed as possible. Therefore we employ the following hyperspherical potential energy $\boldsymbol{E_s}$ as
	\begin{equation}
		\boldsymbol{E_s}\left({\boldsymbol{v}}\right)=\sum_{i=1}^L \sum_{j=1, j \neq i}^L \boldsymbol{e_s}\left(\left\|{\boldsymbol{v}}_i-{\boldsymbol{v}}_j\right\|\right)=\left\{\begin{array}{l}
			\sum_{i \neq j}\left\|{\boldsymbol{v}}_i-{\boldsymbol{v}}_j\right\|^{-s}, \quad s>0 \\
			\sum_{i \neq j} \boldsymbol{\log} \left(\left\|{\boldsymbol{v}}_i-{\boldsymbol{v}}_j\right\|^{-1}\right), \quad s=0
		\end{array}\right.
	\end{equation}
	
	Here $v$ is sampled feature of each image with point-wise embedding $v_i$ and L is the length of the feature, which is also the number of sampled voxels. We randomly sample from the whole case so that the features can better express the overall representational power of the model. The feature vectors will be more widely dispersed in the unit sphere if the hyperspherical energy (HSE) is lower\cite{chen2021contrastive}. For the dataset with $N$ cases, we choose $s=1$ and the feature variety $F_v$ is formulated as 
	
	\begin{equation}
		\boldsymbol{F_v}=\frac{1}{N} \sum_{i=1}^N \boldsymbol{E^{-1}_s(v)}
	\end{equation}

	\noindent\textbf{Overall Estimation.} As for semantic segmentation problems, the feature pyramid structure is critical for segmentation results\cite{lin2017feature}\cite{zhao2017pyramid}. Hence in our framework, different decoders' outputs are upsampled to the size of the output and can be used in the sliding window sampling process. Besides, we decrease the sampling ratio in the decoder layer close to the bottleneck to avoid feature redundancy. The final transferability of pre-trained model $m$ to dataset $t$  $\mathcal{T}_{m\rightarrow t}$ is
	\begin{equation}
		\mathcal{T}_{m\rightarrow t} = \frac{1}{D}\sum_{i=1}^{D}\log\frac{\boldsymbol{F_v^i}}{\boldsymbol{C_{cons}^i}}
	\end{equation}
	where D is the number of decoder layers used in the estimation.
	
	\section{Experiment}
	
	\subsection{Experiment on MSD Dataset}
	
	The Medical Segmentation Decathlon (MSD)~\cite{antonelli2022medical} dataset is composed of ten different datasets with various challenging characteristics, which are widely used in the medical image analysis field. To evaluate the effectiveness of CC-FV, we conduct extensive experiments on 5 of the MSD dataset, including Task03 Liver(liver and tumor segmentation), Task06 Lung(lung nodule segmentation), Task07 Pancreas(pancreas and pancreas tumor segmentation), Task09 Spleen(spleen segmentation), and Task10 Colon(colon cancer segmentation). All of the datasets are 3D CT images. The public part of the MSD dataset is chosen for our experiments, and each dataset is divided into a training set and a test set at a scale of 80\% and 20\%. For each dataset, we use the other four datasets to pre-train the model and fine-tune the model on this dataset to evaluate the performance as well as the transferability using the correlation between two ranking sequences of upstream pre-trained models. We load all the pre-trained models' parameters except for the last convolutional layer and no parameters are frozen during the fine-tuning process. On top of that, we follow the nnUNet\cite{isensee2021nnu} with the self-configuring method to choose the pre-processing, training, and post-processing strategy. For fair comparisons, the baseline methods including TransRate~\cite{huang2022frustratingly}, LogME~\cite{you2021logme}, GBC~\cite{pandy2022transferability} and LEEP~\cite{nguyen2020leep} are also implemented. For these currently available methods, we employ the output of the layer before the final convolution as the feature map and sample it through the same sliding window as CC-FV to obtain different classes of features, which can be used for the calculation.
	
\begin{figure*}[!t]
		\centering    \includegraphics[width=1.0\linewidth]{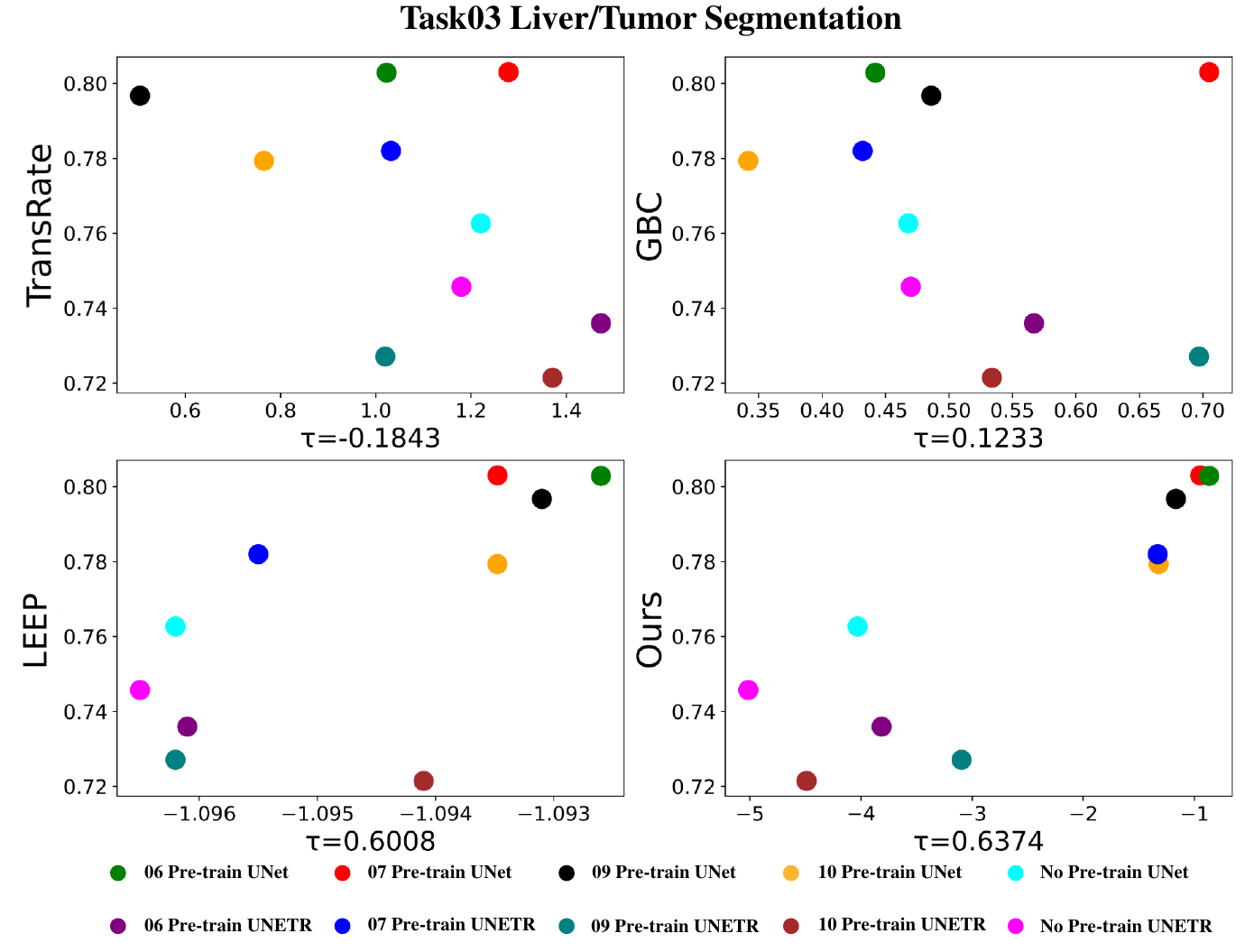}
		\caption{Correlation between the fine-tuning performance and transferability metrics using Task03 as an example. The vertical axis represents the average Dice of the model, while the horizontal axis represents the transferability metric results. We have standardized the various metrics uniformly, aiming to observe a positive relationship between higher performance and higher transferability estimations. }
		\label{fig:rank_result}
	\end{figure*}

	Fig.\ref{fig:rank_result} visualizes the average Dice score and the estimation value on Task 03 Liver. The TE results are obtained from the training set only. U-Net\cite{ronneberger2015u} and UNETR\cite{hatamizadeh2022unetr} are applied in the experiment and each model is pre-trained for 250k iterations and fine-tuned for 100k iterations with batch size 2 on a single NVIDIA A100 GPU. Besides, we use the model at the end of training for inference and calculate the final DSC performance on the test set. And we use weighted Kendall's $\tau$\cite{you2021logme} and Pearson correlation coefficient for the correlation between the TE results and fine-tuning performance. The Kendall's $\tau$ ranges from [-1, 1], and $\tau$=1 means the rank of TE results and performance are perfectly correlated($\mathcal{T}^{i}_{s\rightarrow t}>\mathcal{T}^{j}_{s\rightarrow t}$ if and only if $\mathcal{P}^{i}_{s\rightarrow t}>\mathcal{P}^{j}_{s\rightarrow t}$). Since model selection generally picks the top models and ignores the poor performers, we assign a higher weight to the good models in the calculation, known as weighted Kendall's $\tau$. The Pearson coefficient also ranges from [-1, 1], and measures how well the data can be described by a linear equation. The higher the Pearson coefficient, the higher the correlation between the variables. It is clear that the TE results of our method have a more positive correlation with respect to DSC performance. 
	
	Table.\ref{tab::main_table} demonstrates that our method surpasses all the other methods. Most of the existing methods are inferior to ours because they are not designed for segmentation tasks with a serious class imbalance problem. Besides, these methods rely only on single-layer features and do not make good use of the hierarchical structure of the model.
	
	\begin{figure}[!t]
		\centering
		\includegraphics[width=1.0\linewidth]{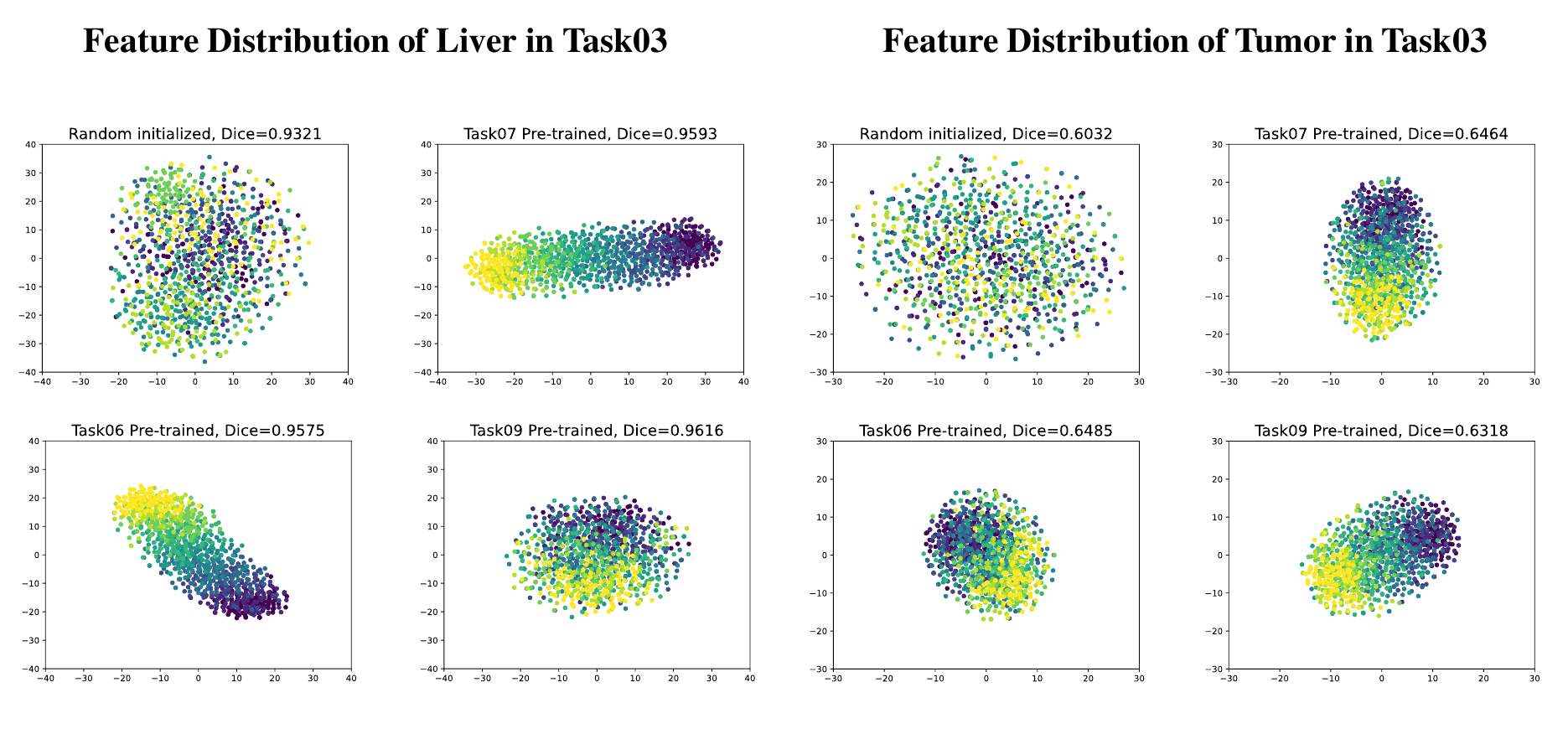}
		\caption{Visualization of features with same labels using t-SNE. Points with different colors are from different samples. Pre-trained models tend to have a more consistent distribution within a class than the randomly initialized model and after fine-tuning they often have a better Dice performance than the randomly initialized models.}
		\label{fig:viz_feat}
	\end{figure}

	\begin{table}[!t]
		\caption{Pearson coefficient and weighted Kendall's $\tau$ for transferability estimation}
		\label{tab::main_table}
		\setlength{\tabcolsep}{2mm}{} 
		\begin{tabular}{c|ccccccl}
			\hline
			Data/Method                & \multicolumn{1}{l}{Metrics} & Task03                      & Task06                      & Task07                     & Task09                     & Task10                      & Avg \\ \hline
			\multirow{2}{*}{LogME}     & $\tau$                         & -0.1628                     & -0.0988                     & 0.3280                     & 0.2778                     & -0.2348                     & 0.0218  \\ \cline{2-8} 
			& pearson                           & \multicolumn{1}{l}{0.0412}  & \multicolumn{1}{l}{0.5713}  & \multicolumn{1}{l}{0.3236} & \multicolumn{1}{l}{0.2725} & \multicolumn{1}{l}{-0.1674} & 0.2082  \\ \hline
			\multirow{2}{*}{TransRate} & $\tau$                         & -0.1843                     & -0.1028                     & 0.5923                     & 0.4322                     & 0.6069                      & 0.2688  \\ \cline{2-8} 
			& pearson                           & \multicolumn{1}{l}{-0.5178} & \multicolumn{1}{l}{-0.2804} & \multicolumn{1}{l}{0.7170} & \multicolumn{1}{l}{0.5573} & \multicolumn{1}{l}{0.7629}  & 0.2478  \\ \hline
			\multirow{2}{*}{LEEP}      & $\tau$                         & 0.6008                      & 0.1658                      & 0.2691                     & 0.3516                     & 0.5841                      & 0.3943  \\ \cline{2-8} 
			& pearson                           & 0.6765                      & \multicolumn{1}{l}{-0.0073} & \multicolumn{1}{l}{0.7146} & \multicolumn{1}{l}{0.1633} & \multicolumn{1}{l}{0.4979}  & 0.4090  \\ \hline
			\multirow{2}{*}{GBC}       & $\tau$                         & 0.1233                      & -0.1569                     & 0.6637                     & 0.7611                     & 0.6643                      & 0.4111  \\ \cline{2-8} 
			& pearson                           & \multicolumn{1}{l}{-0.2634} & \multicolumn{1}{l}{-0.3733} & \multicolumn{1}{l}{0.7948} & \multicolumn{1}{l}{0.7604} & \multicolumn{1}{l}{0.7404}  & 0.3317  \\ \hline
			\multirow{2}{*}{Ours CC-FV}      & $\tau$                         & 0.6374                      & 0.0735                      & 0.6569                     & 0.5700                     & 0.5550                      & \textbf{0.4986}  \\ \cline{2-8} 
			& pearson                           & \multicolumn{1}{l}{0.8608}  & \multicolumn{1}{l}{0.0903}  & \multicolumn{1}{l}{0.9609} & \multicolumn{1}{l}{0.7491} & \multicolumn{1}{l}{0.8406}  & \textbf{0.7003}  \\ \hline
		\end{tabular}
	\end{table}
	
	\subsection{Ablation Study}
	In Table.\ref{tab::ablation} we analyze the different parts of our method and compare some other methods. First, we analyze the impact of class consistency $C_{cons}$ and feature variety $F_v$. Though $F_v$ can not contribute to the final Kendall's $\tau$ directly, $C_{cons}$ with the constraint of $F_v$ promotes the total estimation result. Then we compare the performance of our method at single and multiple scales to prove the effectiveness of our multi-scale strategy. Finally, we change the distance metrics in class consistency estimation. KL-divergence and Bha-distance are unstable in high dimension matrics calculation and the performance is also inferior to the Wasserstein distance. Fig.\ref{fig:viz_feat} visualize the distribution of different classes using t-SNE methods. We can easily find that with models with a pre-training process have a more compact intra-class distance and a higher fine-tuning performance.

	\begin{table}[!t]
		\caption{Ablation on the effectiveness of different parts in our methods}
		\label{tab::ablation}
		\setlength{\tabcolsep}{2.5mm}{} 
		\centering
		\begin{tabular}[]{c|ccccc|c}
			\hline
			Data/Method & Task03 & Task06 & Task07 & Task09 & Task10 &  Avg \\ \hline
			Ours CC-FV      &     0.6374           &      0.0735       &      0.6569     &      0.5700     &    0.5550       &     \textbf{0.4986}     \\ \hline
			Ours w/o $C_{cons}$        &     0.1871   &    -0.2210    &    -0.2810    &     -0.0289   &  -0.2710      &    -0.1230     \\ \hline
			Ours w/o $F_v$       &   0.6165     &   0.3235     &    0.6054    &      0.2761  &   0.5269     &    0.4697     \\ \hline
			Single-scale        &   0.4394     &   0.0252     &   0.5336     &     0.5759   &     0.6007   &    0.4341     \\ \hline
			KL-divergence       &     -0.5658         &      -0.0564       &    0.2319       &      0.4628      &      -0.0323     &      0.0080    \\ \hline
			Bha-distance        &   0.1808     &   0.0723     &    0.2295    &    0.7866    &   0.4650     &    0.3468     \\ \hline

		\end{tabular}
	\end{table}

	\section{Conclusion}
	
	In our work, we raise the problem of model selection for upstream and downstream transfer processes in the medical image segmentation task and analyze the practical implications of this problem. In addition, due to the ethical and privacy issues inherent in medical care and the computational load of 3D image segmentation tasks, we design a generic framework for the task and propose a transferability estimation method based on class consistency with feature variety constraint, which outperforms existing model transferability estimation methods as demonstrated by extensive experiments.

	\section*{Acknowledgement}
	This work was supported by the Open Funding of Zhejiang Laboratory under Grant 2021KH0AB03, NSFC China (No. 62003208); Committee of Science and Technology, Shanghai, China (No.19510711200); Shanghai Sailing Program (20YF1420800), and Shanghai Municipal of Science and Technology Project (Grant No.20JC1419500, No. 20DZ2220400).
	%
	%
	%
	%

		
		
		
	
	\bibliographystyle{splncs04}
	\bibliography{refs}
\end{document}